\documentclass{article}
\usepackage{spconf,amsmath,graphicx}

\usepackage[enable]{easy-todo} % enable or disable
\usepackage{graphicx}          % graphics
\usepackage{lipsum}            % You-know-what spell; e.g., \lipsum[2] is Nam dui ligula, fringilla a, ...
\usepackage{float}             % H instead of htbp (the H float specifier). here package was withdrawn many years ago.
\usepackage{wrapfig}           % wrap figure
\usepackage{xcolor}            % colors
\usepackage{url}               % simple URL typesetting
\usepackage{hyperref}          % hyperlinks
\usepackage{booktabs}          % professional-quality tables
\usepackage{multirow}
\usepackage{subcaption}        % needed for subtables.
\usepackage{amsfonts}          % blackboard math symbols
\usepackage{amsmath, amssymb, amsthm} % align cases, $\mathbb{R}$
\usepackage{mathtools}         % E.g., \xrightarrow[\text{lower}]{\text{$t \rightarrow \infty$}}
\usepackage{nicefrac}          % compact symbols for 1/2, etc.
\usepackage{microtype}         % microtypography
\usepackage{type1cm}           % $\mathbb{R} etc.$
\usepackage{bm}                % \bm{vector}
\usepackage{dsfont}            % \mathds{1} is the indicator function (Use this instead of \mathbbm{1}: bbm package embeds type3 font.)
\usepackage{cases}             % simultaneous equations.
\usepackage{array}             % array environment for matrix etc.
\usepackage{fancyhdr}

\usepackage{caption}
\usepackage{xr}

\title{Rethinking the Backbone in Class Imbalanced Federated Source Free Domain Adaptation: The Utility of Vision Foundation Models}
\name{Kosuke Kihara, Junki Mori, Taiki Miyagawa, Akinori F. Ebihara}
\address{NEC Corporation, Kawasaki, Japan}
\fancypagestyle{firstpage}{
  \fancyhf{}                  % clear all header/footer fields

  \fancyfoot[C]{%
  \fontsize{6.73}{8}\selectfont
    \copyright~2025 IEEE.  Personal use of this material is permitted.  Permission from IEEE must be obtained for all other uses, in any current or future media, including reprinting/republishing this material for advertising or promotional purposes, creating new collective works, for resale or redistribution to servers or lists, or reuse of any copyrighted component of this work in other works.
  }%
}

\begin{document}

\thispagestyle{firstpage}

\maketitle

\begin{abstract} % maximum of 200 words
Federated Learning (FL) offers a framework for training models collaboratively while preserving data privacy of each client. Recently, research has focused on Federated Source-Free Domain Adaptation (FFREEDA), a more realistic scenario wherein client-held target domain data remains unlabeled, and the server can access source domain data only during pre-training. We extend this framework to a more complex and realistic setting: Class Imbalanced FFREEDA (CI-FFREEDA), which takes into account class imbalances in both the source and target domains, as well as label shifts between source and target and among target clients. The replication of existing methods in our experimental setup lead us to rethink the focus from enhancing aggregation and domain adaptation methods to improving the feature extractors within the network itself. We propose replacing the FFREEDA backbone with a frozen vision foundation model (VFM), thereby improving overall accuracy without extensive parameter tuning and reducing computational and communication costs in federated learning. Our experimental results demonstrate that VFMs effectively mitigate the effects of domain gaps, class imbalances, and even non-IID-ness among target clients, suggesting that strong feature extractors, not complex adaptation or FL methods, are key to success in the real-world FL.

\end{abstract}
\begin{keywords}
Domain adaptation, Source-free Domain Adaptation, Federated Learning, Class Imbalance, Foundation Model
\end{keywords}
%

% ================================================================== %
\section{Introduction} \label{sec:intro}
\begin{figure}[t]
    \centering
    \includegraphics[width=0.48\textwidth]{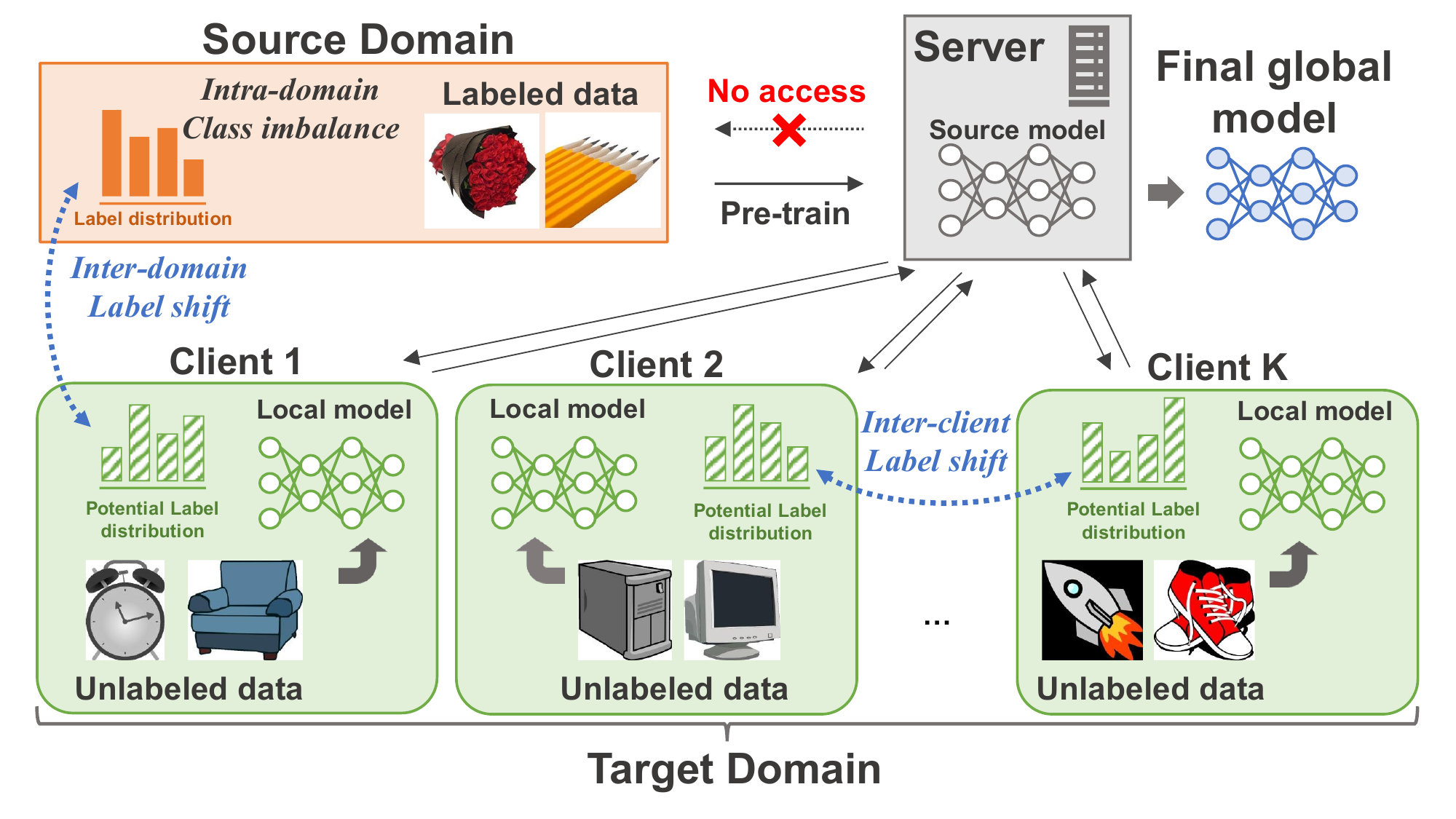}
    \caption{Overview of our setting: \textbf{CI-FFREEDA}. Federated target adaptation is conducted using a pre-trained model trained from labeled data in the source domain. Clients possess unlabeled data from a single target domain and do not have access to the source domain. The goal is to acquire a global model adapted to the target domain. 
    There is an imbalance within each domain and client (\textit{intra-domain class imbalance}), resulting in a label distribution gap between the source and target domains (\textit{inter-domain label shift}), as well as among clients (\textit{inter-client label shift}).}
    \label{fig:overview}
\end{figure}

% ================================================================== %
\indent
Modern federated learning (FL) promises to train powerful models while keeping raw data on user devices, an attractive property for privacy-sensitive domains such as mobile health and smart homes.
However, real-world deployments rarely match the clean laboratory conditions assumed in much of the literature.
Three major challenges must be overcome:
(i) \textbf{Unsupervised Domain Adaptation}: A gap often exists between the domain of the pre-trained model held by the server (source domain) and the domain of the data collected by clients (target domain), necessitating solutions for domain adaptation. Due to the lack of specialists and the high cost of annotation, this adaptation often operates in an unlabeled manner.
(ii) \textbf{Source-Free unsupervised Domain Adaptation (SFDA)}: Legal or practical constraints frequently prohibit access to the labeled source domain data once the target adaptation begins. This restriction prevents the fine-tuning routines that dominate conventional domain adaptation pipelines.
(iii) \textbf{Severe class imbalance}: 
The source dataset is often class-imbalanced.
Additionally, the client dataset can also suffer from the inter-client imbalance, with differing majority classes across clients. This compounding of imbalances leads to three levels: intra-domain class imbalance within each domain, inter-domain label shifts (imbalance ratio shifts) between the source and target, and inter-client label shifts among target clients (Fig.~\ref{fig:overview}).

We focus on these three challenges to simulate more practical and realistic scenarios than prior studies in FL, which have only addressed challenges (i) and (ii), known as Federated source-Free Domain Adaptation (FFREEDA) \cite{shenaj2023ladd, mori2025fedwca}.
We extend FFREEDA to class-imbalanced settings, introducing \textbf{CI-FFREEDA}.
By combining representative FL and SFDA methods, we conduct a systematic evaluation. 
The results show that existing SFDA methods, including those explicitly designed for class-imbalanced settings, struggle to demonstrate their effectiveness.
This highlights the need to address the underlying challenges beyond refining FL and SFDA methods.
Guided by these findings, we revisit the significance of the backbone model in CI-FFREEDA, which should extract domain-invariant features. 
Specifically, we propose employing a feature extractor derived from vision foundation models (VFMs) pre-trained on large-scale datasets. To maximize their feature extracting capabilities, we suggest freezing the parameters of VFMs and fine-tuning only the subsequent parameters. This approach reduces memory consumption compared to previous studies, which typically fine-tune ImageNet pre-trained models as the feature extractor. Furthermore, since the frozen VFM can be shared among all target clients, it is not necessary to communicate the feature extractor during model aggregation.

This paper makes three contributions:
\begin{enumerate}
\item In CI-FFREEDA, we propose an approach that adopts VFMs as the model architecture while reducing computational and communication costs in federated learning.
\item We conduct experiments combined with existing FL and 
 SFDA methods to demonstrate that the performance of many methods can be improved by using VFMs.
\item Through detailed experiments, we further show that VFMs bridge domain gaps and label distribution gaps, serving as a link between federated learning, source-free domain adaptation, and class imbalance problems.

\end{enumerate}

% ================================================================== %
\section{Related Work} \label{sec:related_work}
% ================================================================== %

SFDA is actively studied as a privacy-conscious approach to domain adaptation. One of the most representative method is SHOT~\cite{liang2020ashot}, which combines information maximization with self-training through pseudo-labeling based on prototypes. NRC~\cite{yanq2021nrc} and AaD~\cite{yanq2022aad} assume consistency of predictions among neighboring samples. Other approaches include 3C-GAN~\cite{rui20203cgan}, which uses generative models to produce source domain-like samples, and A$^2$Net~\cite{haifeng2021a2net}, which constructs a target-specific classifier through adversarial inference.
ISFDA~\cite{li2021isfda} is the first study to address intra-domain class imbalance and inter-domain label shift, modifying pseudo-labels using the certainty of inference results. In the same imbalanced conditions, ICPR~\cite{tian2025icpr} conducts training to ensure that predictions after different augmentations of reliable samples are consistent with those of neighboring samples.
Recently, there has been increasing interest in enhancing pseudo-label accuracy using vision-language models \cite{song2024difo, wenyu2024colearn}. However, these approaches assume a highly resource-rich target adaptation environment.

FedAvg~\cite{mcmahan2017fedavg} is proposed as a framework for collaboratively constructing superior models while preserving privacy, even in situations lacking abundant computing environments and data resources. Subsequent research has progressed on methods effective under non-independent and identically distributed (non-IID) conditions among clients~\cite{li2020fedprox}. Existing studies that integrate domain adaptation with FL~\cite{peng2020fada, yao2022dualadapt, gan2022fruda}, 
typically assume the access to source domain data. Efforts to achieve federated source-free domain adaptation (FFREEDA) are still in their early stages \cite{shenaj2023ladd, mori2025fedwca}, and the potential non-IID-ness between clients that may arise in practice has not yet been thoroughly examined.

% ================================================================== %
\section{Methodology} \label{sec:method}
% ================================================================== %
\subsection{Problem formulation} 
\label{subsec:problem}

We focus on federated SFDA in the presence of class imbalance (CI-FFREEDA). The server trains a source model using labeled data from a source domain, denoted as $D_s = \{(x_s^i, y_s^i)\ |\ i=1,\ldots,N_s\}$. This trained source model is distributed to each of the $K$ clients (indexed by $k$), for training on unlabeled target domain data, represented as  $D_t^{(k)} = \{x_t^{(k),i}\ |\ i=1,...,N_t^{(k)}\}$. 
In this paper, both the source and target domains are single, i.e., a scenario where adaptation occurs from a single source domain to a single target domain across multiple clients.
The final goal is to develop the best possible single global model that has been adapted to the target domain. 
Typically, the assumption of class imbalanced domain adaptation \cite{li2021isfda} includes the joint occurrence of covariate shift and label shift, expressed as:
$p(x) \neq q(x)$, $p(y|x)=q(y|x)$, $p(y)\neq q(y)$, and $p(x|y)\neq q(x|y)$. Here, $p(\cdot)$ and $q(\cdot)$ represent the distribution of source and target domains, respectively. 
In our FL setup, the range of covariate shift and label shift is extended to across the source domain and all target clients:
$p(x_s) \neq q(x_t^{1}) \neq \ldots \neq q(x_t^{K})$, $p(y_s) \neq q(y_t^{1}) \neq \ldots \neq q(y_t^{K})$, and $p(y_s | x_s) = q(y_t^{1} | x_t^{1}) = \ldots = q(y_t^{K} | x_t^{K})$.
Additionally, we also account for the domain shift between the source and target:
$p(x_s | y_s) \neq q(x_t^{k} | y_t^{k})$ and $q(x_t^{1} | y_t^{1}) = \ldots = q(x_t^{K} | y_t^{K})$.
\subsection{Proposed Method}

The method we propose in CI-FFREEDA is as follows:
(i) Substitute the backbone of the existing SFDA framework from a traditional CNN-based network to a VFM. 
(ii) Freeze the parameters of the backbone during both the source training phase and the federated target adaptation phase, focusing on training and aggregating the bottleneck and classifier components. 
(iii) Implement balanced sampling to uniformly distribute the number of samples per class within each batch during the source training phase.

We select DINOv2~\cite{oquab2024DINOv2} as the VFM. DINOv2 is the Vision Transformer (ViT) pre-trained on 142 million images~\cite{dosovitskiy2021vit}, named ViT-giant, which has 1.1 billion parameters, and several distilled models are available.
In this paper, we employ the ViT-S/14 distilled model (21M parameters) and ViT-B/14 distilled model (86M parameters), and they are compared with ResNet-50~\cite{he2016resnet} ($\sim$25M) and ResNet-101 ($\sim$45M) pretrained on ImageNet.
The advanced feature extraction capability of VFMs is expected to achieve universal extraction that is invariant to class imbalance and domain differences.

Some previous studies adopt the hypothesis transfer from SHOT \cite{liang2020ashot} and freeze the source classifier during target adaptation. However, our proposed approach addresses the limitation in the expressive capacity of bottleneck due to the fixed backbone and the variability of imbalance ratios across different clients, thus we set the classifier trainable.

\begin{figure}[t]
    \centering
    \includegraphics[width=0.48\textwidth]{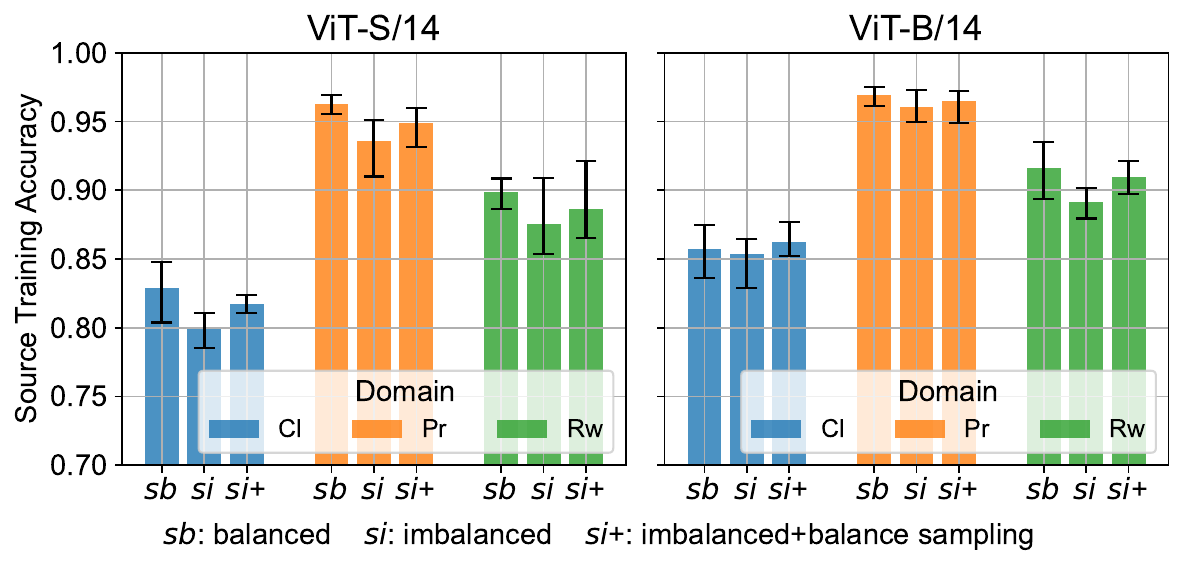}
    \caption{VFMs with balance sampling in source training reduce imbalance effects. The figure shows three types of training: source model trained on balanced data (\textit{sb}), imbalanced data (\textit{si}), and imbalanced data with balance sampling (\textit{si+}). DINOv2 ViT-S and ViT-B are used as the VFMs. The dataset is Office-Home. The performance metric is the macro-averaged recall. Colors indicate different source domains. Each plot is the average over nine runs, comprised of three different source sampling and three execution seeds for each source sampling. Error bars represent the maximum and minimum values of 9 runs.}
    \label{fig:srcoh}
\end{figure}

Fig.~\ref{fig:srcoh} shows the macro averaged recall (see Sec.~\ref{subsec:setup}) of source models using DINOv2 ViT-S and ViT-B with Office-Home~\cite{venkateswara2017oh}.
The accuracy trained with imbalanced data (\textit{si}) is lower than that trained with balanced data (\textit{sb}); however, balanced sampling (\textit{si+}) significantly mitigates this discrepancy. 
We finally note that freezing the high-performance VFM backbone significantly reduces training time, memory consumption, and communication costs in FL (details are shown in \href{https://sigport.org/documents/appendix-rethinking-backbone-class-imbalanced-federated-source-free-domain-adaptation}{supplementary material}\footnote{\href{https://sigport.org/documents/appendix-rethinking-backbone-class-imbalanced-federated-source-free-domain-adaptation}{https://sigport.org/documents/appendix-rethinking-backbone-class-imbalanced-federated-source-free-domain-adaptation}}), and also simplifies hyperparameter tuning compared to training a backbone from scratch or fine-tuning an ImageNet pre-trained model.

% ================================================================== %
\section{Experiments} \label{sec:experiments}
% ================================================================== %

\subsection{Experimental Setup}
\label{subsec:setup}

\textbf{Datasets.}
We use two datasets for domain adaptation.
(i)~\textbf{Office-Home (OH)}~\cite{venkateswara2017oh} is a medium-sized benchmark for domain adaptation. It comprises four domains and 65 classes: Art (2,427 images), Clipart (\textbf{Cl}; 4,365 images), Product (\textbf{Pr}; 4,439 images), and Real-World (\textbf{Rw}; 4,357 images). Due to the limited sample size in the Art domain, we employ the remaining three domains.
(ii) \textbf{VisDA-C (VisDA)}~\cite{peng2017visda} is a large benchmark consisting of a 3D rendering source domain (152k images) and a photo-realistic target domain (55k images) across 12 classes.

\noindent
\textbf{Models.} 
Our model architecture is based on existing SFDA methods such as SHOT \cite{liang2020ashot}. It consists of a feature extractor, a bottleneck for dimension adjustment, and a classifier. We use DINOv2 ViT-S and ViT-B as the backbone for the feature extractor. Following prior studies, we prepare ResNet-50~\cite{he2016resnet} for OH and ResNet-101 for VisDA as backbones for comparison.

\noindent
\textbf{Evaluation.} 
We prepare imbalanced datasets for training and validation from source domain data. Target domain data is partitioned into imbalanced training set, imbalanced validation set, and balanced test set. The training and validation set are distributed to each target client with Dirichlet random sampling.
The training validation set is employed to assess the effectiveness of imbalance learning for each client. We use macro averaged recall (MAR) as the evaluation metric, which is the average of the recall for each class across all classes. In this paper, the term ``accuracy'' specifically refers to MAR. In order to obtain a unified global model adapted to the target domain data, we maintain the best global model on the server based on the evaluations from each client. The final evaluation of the global model is conducted using the balanced test set from the administrator's perspective.

\noindent
\textbf{Methods.}
We choose three standard SFDA methods with low learning costs for clients (SHOT \cite{liang2020ashot}, NRC \cite{yanq2021nrc}, and AaD \cite{yanq2022aad}) and two SFDA methods designed for the class imbalanced setting (ISFDA \cite{li2021isfda} and ICPR \cite{tian2025icpr}).
We adopt FedAvg \cite{mcmahan2017fedavg} for the federated learning algorithm\footnote{LADD~\cite{shenaj2023ladd} and FedWCA~\cite{mori2025fedwca} are not included as methods because they are designed for multi-target adaptation, while we focus on single-target adaptation in this paper.
}.

\noindent
\textbf{Implementation Details.}
The number of target clients ($K$) is set to 3 for OH and a more challenging setting of $K = 50$ for VisDA.
The number of epochs in source training is set to 40 in OH and 10 in VisDA, and the number of local epochs is set to 5 and the communication round to 10 (but in some experiments, it is extended to 20) in target adaptation. 
The learning schemes and hyperparameters follow that of previous studies, but in the experiments with DINOv2, the classifier is trainable for all methods. When fine-tuning the ResNet backbone, the classifier's training status follows the respective method.
Detailed setup in our experiments are provided in the \href{https://sigport.org/documents/appendix-rethinking-backbone-class-imbalanced-federated-source-free-domain-adaptation}{supplementary material}\footnotemark[1].

\subsection{Results} \label{subsec:results}

\begin{figure}[h]
    \centering
    \includegraphics[width=0.48\textwidth]{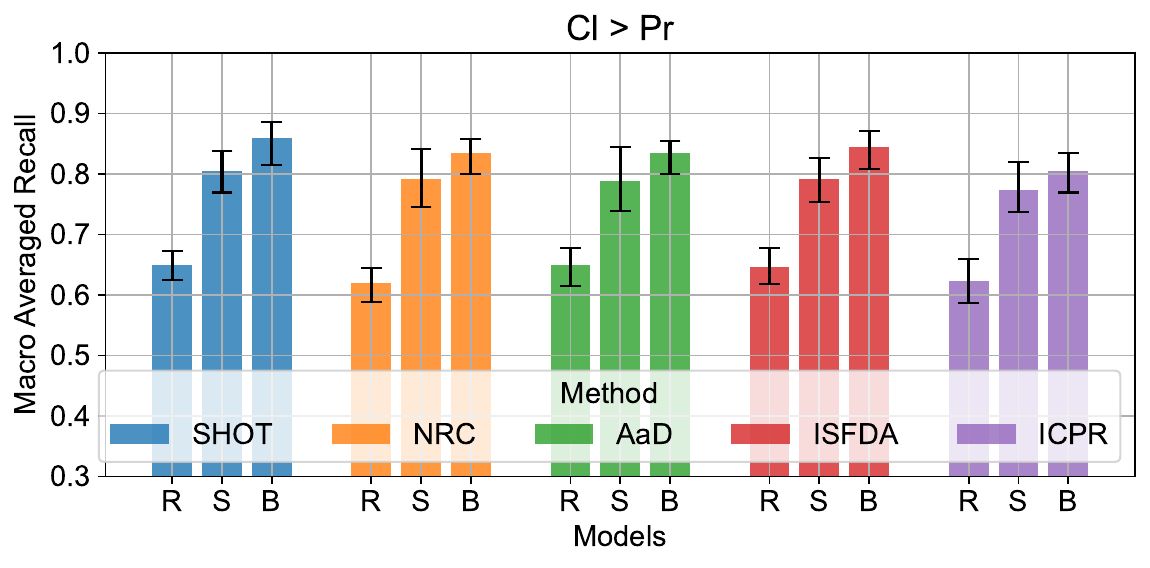}
    \caption{Frozen VFMs enhance SFDA methods.
    It shows the result of adaptation from Clipart to Product in OH with ResNet-50 (R), DINOv2 ViT-S (S), and ViT-B (B). Colors indicate different SFDA methods. Each plot is the average over nine runs, comprised of three different source imbalance ratios and three different target imbalance ratios. Error bars represent the maximum and minimum of nine runs. }
    \label{fig:tgtoh_c2p}
\end{figure}

Fig.~\ref{fig:tgtoh_c2p} illustrates the increase in accuracy after adaptation achieved by using VFMs as the backbone instead of ResNet-50. Here, Fig.~\ref{fig:tgtoh_c2p} shows the result of adaptation from Clipart to Product on the OH dataset by changing the model architecture and SFDA method. The improvements follow the order of ResNet-50 (R), ViT-S (S), and ViT-B (B), with none of the methods exceeding the range of statistical error of the others. Averaging across the six source-target domain pairs in the OH dataset, replacing ResNet-50 with ViT-S improves accuracy by 10.5\%, and by 16.6\% with ViT-B, as indicated in Table~\ref{tab:tgt}.

\begin{figure}[h]
    \centering
    \includegraphics[width=0.48\textwidth]{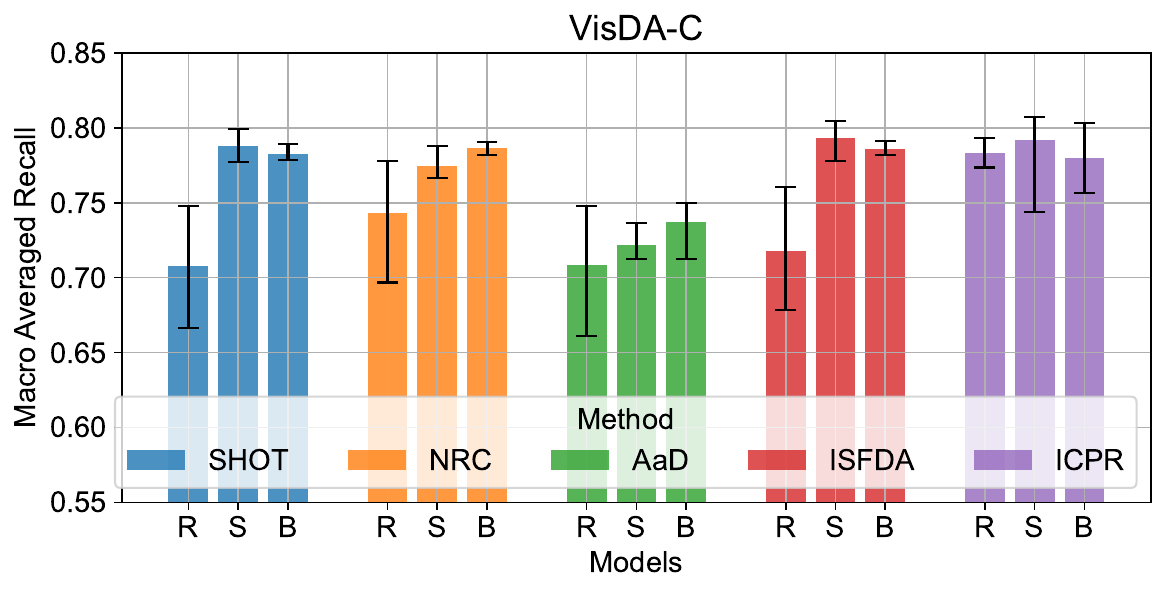}
    \caption{Results on VisDA with ResNet-101 (R), DINOv2 ViT-S (S), and ViT-B (B). Details are the same as Fig.~\ref{fig:tgtoh_c2p}.}
    \label{fig:tgtvisda}
\end{figure}

The advantage of VFMs in VisDA is also indicated by Fig.~\ref{fig:tgtvisda}.
Similar to the results observed in OH, most methods exhibit low accuracy and significant variation for ResNet-101. 
A key difference from OH is the performance of ICPR, the most recent method, achieves relatively better accuracy compared to the others when using ResNet-101. However, this advantage diminishes when using VFMs, where ICPR achieves comparable accuracy to other methods and exhibits greater variation. This observation suggests that the robust feature extraction capability of VFMs mitigates the variation induced by the data augmentation used in ICPR.
Another observation is that the improvement from ViT-S to ViT-B is less substantial compared to that in OH. Although both models show an average improvement of 4.2\%, there is no significant difference in accuracy between them (Table~\ref{tab:tgt}). Given the reduction in the number of categories from 65 in OH to 12 in VisDA, it is possible that a large feature extractor such as ViT-B may not be requisite.

\begin{table}
    \centering
    \caption{Summary of Sec.~\ref{sec:experiments}, showing that frozen VFMs enhance SFDA methods. Each accuracy is averaged over all runs and source-target domain pairs. The average accuracy across all methods and the difference from ResNet (ResNet-50 for OH and ResNet-101 for VisDA) results are also shown.}
    \label{tab:tgt}
    \scalebox{0.74}{
    \begin{tabular}{cc|ccccc|cc}
    \toprule
    Data & Model & SHOT & NRC & AaD & ISFDA & ICPR & Ave. & diff. \\
    \midrule
            & ResNet & 64.7 & 61.9 & 63.7 & 64.1 & 61.8 & 63.2 & -- \\
    OH & ViT-S & 75.1 & 74.2 & 71.8 & 74.2 & 73.2 & 73.7 & +10.5 \\
                & ViT-B & 81.2 & 80.0 & 80.5 & 79.8 & 77.6 & 79.8 & +16.6 \\
    \midrule
       & ResNet & 70.7 & 74.3 & 70.8 & 71.8 & 78.3 & 73.2 & -- \\
    VisDA & ViT-S & 78.8 & 77.5 & 72.2 & 79.3 & 79.2 & 77.4 & +4.2 \\
            & ViT-B & 78.3 & 78.7 & 73.7 & 78.6 & 78.0 & 77.4 & +4.2 \\
    \bottomrule 
    \end{tabular}
    }
\end{table}

% ================================================================== %
\section{Robustness of VFMs to Data Gaps}
% ================================================================== %
\subsection{Domain Gaps and Class Imbalances}
\label{subsec:further_gap}
\noindent
\textbf{Setup.}
To investigate the behavior of frozen VFMs in the presence of several data gaps, we perform a series of experiments alternating conditions with and without a domain gap and class imbalances. These experiments are conducted with FedAvg and SHOT in OH.
For the case of no domain gap, transfer learning (TL) setting where source and target domains are the same, the samples from each domain are divided into source and target sets. In the case of heterogeneous domain adaptation (DA) setting, a source set from one domain and a target set from another domain are used. Two configurations for dividing samples into source and target sets are used: balanced and imbalanced. The target set is distributed among three clients, as described in Sec.~\ref{subsec:setup}.
The experiments in two label distribution scenario are shown: source balanced to target balanced (\textit{sbtb}) and source imbalanced to target imbalanced (\textit{siti}). The results of other two scenarios (\textit{sbti} and \textit{sitb}) are presented in the \href{https://sigport.org/documents/appendix-rethinking-backbone-class-imbalanced-federated-source-free-domain-adaptation}{supplementary material}\footnotemark[1].

\vskip.5\baselineskip
\noindent
\textbf{Domain Gaps.}
The effectiveness of VFMs for domain adaptation under FL is indicated by Table~\ref{tab:further_da}, which is an experiment in DA setting and \textit{sbtb} scenario. It shows that both source and target accuracies are higher when using both ViT models compared to ResNet-50. Moreover, ViT models exhibit lower accuracy degradation due to target adaptation compared to ResNet-50, with 17.6\% for ResNet-50, 10.6\% for ViT-S, and 7.3\% for ViT-B. This indicates that both ViT models can extract domain invariant features and bridge domain gaps.

\begin{table}[ht]
    \centering
    \caption{Experiments under DA setting and \textit{sbtb} scenario, showing the decline after the adaptation to the target (S2T diff.). 
    Six domains and nine runs are averaged.
    }
    \label{tab:further_da}
    \scalebox{0.75}{
    \begin{tabular}{cc|c|ccc}
    \toprule
     Setting & Model  & Scenario  & Source & Target & S2T   \\
             &           &           & acc.   & acc.   & diff. \\
    \midrule
        & ResNet-50 & \textit{sbtb} & 82.6 & 65.0 & \textbf{-17.6} \\ 
     DA & ViT-S     & \textit{sbtb} & 87.4 & 76.8 & \textbf{-10.6} \\ 
        & ViT-B     & \textit{sbtb} & 89.9 & 82.6 & \textbf{-7.3} \\
    \bottomrule 
    \end{tabular}
    }
\end{table}

\vskip\baselineskip
\noindent
\textbf{Intra-domain class imbalance.}
Table~\ref{tab:further_idci} illustrates that VFMs effectively mitigate the effects of imbalance. Here, Table~\ref{tab:further_idci} compares two scenarios in TL setting. 
The accuracy degradation after target adaptation between \textit{sbtb} ($^{*}$) and \textit{siti} ($^{\dagger}$) is 4.4\% for ResNet-50, 3.0\% for ViT-S, and 2.9\% for ViT-B. 
These results indicate that VFMs experience less accuracy degradation compared to ResNet, highlighting their superior performance. It suggests that VFMs are more effective in addressing the impact of data imbalance during source training and target
adaptation. Consequently, ViTs can manage intra-domain class imbalance more efficiently.

\begin{table}[ht]
    \centering
    \caption{The scenarios of \textit{sbtb} and \textit{siti} under TL setting. The difference of target accuracy between \textit{sbtb} ($^{*}$) and \textit{siti} ($^{\dagger}$) (Target diff.) is shown. Three domains and nine runs for each domain are averaged.}
    \label{tab:further_idci}
    \scalebox{0.75}{
    \begin{tabular}{cc|c|cccc}
    \toprule
     Setting & Model     & Scenario & Source & Target & Target diff. \\
             &           &          & acc.   & acc.   & ($^{\dagger}$ - $^{*}$) \\
    \midrule
    \multirow{6}{*}{TL} & \multirow{2}{*}{ResNet-50} & \textit{sbtb}  & 82.6 & 82.0$^{*}$ & \multirow{2}{*}{\textbf{-4.4}}    \\
      &                        & \textit{siti} & 78.7 & 77.6$^{\dagger}$ &   \\ \cline{2-6}
      & \multirow{2}{*}{ViT-S} & \textit{sbtb}     & 87.4 & 86.4$^{*}$ & \multirow{2}{*}{\textbf{-3.0}}    \\ 
      &                        & \textit{siti} & 84.7 & 83.4$^{\dagger}$ &   \\ \cline{2-6}
      & \multirow{2}{*}{ViT-B} & \textit{sbtb}     & 89.9 & 90.1$^{*}$ & \multirow{2}{*}{\textbf{-2.9}}    \\ 
      &                        & \textit{siti} & 88.8 & 87.2$^{\dagger}$ &   \\
    \bottomrule 
    \end{tabular}
    }
\end{table}
\noindent
\textbf{Inter-domain label shift.}
Table~\ref{tab:further_idls} suggests VFM's capability to handle label shifts between domains. The results for \textit{sbtb} and \textit{siti} in DA setting reveal that the degradation from source accuracy in the absence of imbalance (\textit{sbtb} $^{*}$) to target accuracy when both source and target domains exhibit imbalance (\textit{siti} $^{\dagger}$) is 21.8\% for ResNet-50, 16.0\% for ViT-S, and 11.8\% for ViT-B. Although the effects of managing class imbalance within each domain cannot be entirely isolated from these results, they imply that VFMs remain effective  even in scenarios involving inter-domain label shifts.

\begin{table}[ht]
    \centering
    \caption{The scenarios of \textit{sbtb} and \textit{siti} under DA setting. Source-to-target difference under label shift, the difference from source accuracy in \textit{sbtb} ($^{*}$) to target accuracy in \textit{siti} ($^{\dagger}$) (S2T diff. under LS) is shown. Six domain patterns and nine runs for each pattern are averaged.}
    \label{tab:further_idls}
    \scalebox{0.75}{
    \begin{tabular}{cc|c|cccc}
    \toprule
     Setting & Model     & Scenario & Source & Target & S2T diff. under LS\\
             &           &          & acc.   & acc.   & ($^{\dagger}$ - $^{*}$) \\
    \midrule
    \multirow{6}{*}{DA}  & \multirow{2}{*}{ResNet-50} & \textit{sbtb} & 82.6$^{*}$ & 65.0 & \multirow{2}{*}{\textbf{-21.8}}    \\
       &                        & \textit{siti} & 78.7 & 60.8$^{\dagger}$ &  \\ \cline{2-6}
       & \multirow{2}{*}{ViT-S} & \textit{sbtb} & 87.4$^{*}$ & 76.8 & \multirow{2}{*}{\textbf{-16.0}}    \\
       &                        & \textit{siti} & 84.7 & 71.4$^{\dagger}$ &  \\ \cline{2-6}
       & \multirow{2}{*}{ViT-B} & \textit{sbtb} & 89.9$^{*}$ & 82.6 & \multirow{2}{*}{\textbf{-11.8}}    \\
       &                        & \textit{siti} & 88.8 & 78.1$^{\dagger}$ &  \\
    \bottomrule 
    \end{tabular}
    }
\end{table}

\subsection{Label Shifts among Clients}
\label{subsec:further_fed}
\noindent
\textbf{Setup.}
To evaluate the performance of VFMs under label shifts among clients, we conduct the experiments of two non-IID FL methods, FedProx~\cite{li2020fedprox} and FedETF~\cite{li2023fedetf}. % in Table~\ref{tab:further_icls}.
FedProx~\cite{li2020fedprox} adds the differences of model updates as a regularization and enhances the robustness against data heterogeneity. FedETF~\cite{li2023fedetf} builds upon the theory of neural collapse~\cite{vardan2020nc}, which posits that the class prototypes establish an Equiangular Tight Frame (ETF) at the terminal phase of training, presumably mitigating the imbalance bias.
The experiments are conducted under the same setup described in Sec.~\ref{subsec:setup} using SHOT.

\vskip.5\baselineskip
\noindent
\textbf{Inter-client label shift.}
Table~\ref{tab:further_icls} shows that non-IID FL methods exhibit a decline or no significant difference in most cases, and no method is observed to be consistently effective. 
The only exception is observed with ResNet-101 in VisDA, where FedETF mitigates the non-IID-ness using ResNet.
These results suggest that utilizing a frozen VFM as a shared backbone across target clients in FedAvg can mitigate the non-IID-ness of target clients without relying on non-IID FL methods.

\begin{table}[ht]
    \centering
    \caption{Comparison of different FL methods: FedAvg, FedProx, and FedETF. Each accuracy is averaged over nine runs.}
    \label{tab:further_icls}
    \scalebox{0.75}{
    \begin{tabular}{ccc|ccc}
    \toprule
    Data & Model & Domain & FedAvg & FedProx & FedETF \\
    \midrule

OH  & ResNet-50 & Cl$\rightarrow$Pr & 64.9 & 65.1  & \textbf{66.2} \\
    &       & Cl$\rightarrow$Rw & 67.9 & \textbf{68.5}  & \textbf{68.5} \\
    &       & Pr$\rightarrow$Cl & \textbf{48.8} & 47.8  & 47.4 \\
    &       & Pr$\rightarrow$Rw & \textbf{75.5} & 75.1  & 74.7 \\
    &       & Rw$\rightarrow$Cl & \textbf{53.3} & 52.6  & 51.9 \\
    &       & Rw$\rightarrow$Pr & \textbf{77.6} & 76.9  & 76.8 \\ \cline{3-6}
    &       & Ave.              & 64.7 & 64.3           & 64.3 \\ \cline{2-6}
    & ViT-S & Cl$\rightarrow$Pr & \textbf{80.5} & 80.4  & 79.4 \\
    &       & Cl$\rightarrow$Rw & \textbf{81.3} & 80.4  & 80.6 \\
    &       & Pr$\rightarrow$Cl & 59.0 & 59.7  & \textbf{61.4} \\
    &       & Pr$\rightarrow$Rw & 82.8 & \textbf{83.5}  & 81.9 \\
    &       & Rw$\rightarrow$Cl & 61.9 & 62.7  & \textbf{65.1} \\
    &       & Rw$\rightarrow$Pr & 85.2 & \textbf{85.7}  & 84.9 \\ \cline{3-6}
    &       & Ave.              & 75.1 & 75.4           & 75.6 \\ \cline{2-6}
    & ViT-B & Cl$\rightarrow$Pr & 85.9 & \textbf{86.1}  & 85.1 \\
    &       & Cl$\rightarrow$Rw & 84.2 & \textbf{84.3}  & 83.1 \\
    &       & Pr$\rightarrow$Cl & 69.1 & 69.3  & \textbf{70.8} \\
    &       & Pr$\rightarrow$Rw & 86.5 & \textbf{86.7}  & 85.5 \\
    &       & Rw$\rightarrow$Cl & 71.9 & 72.3  & \textbf{73.6} \\
    &       & Rw$\rightarrow$Pr & 89.7 & \textbf{89.9}  & 89.6 \\ \cline{3-6}
    &       & Ave.              & 81.2 & 81.4           & 81.3 \\

    \midrule
     VisDA & ResNet-101 & -- & 70.7 & 70.8 & \textbf{75.3} \\
           & ViT-S & -- & 78.8 & 78.7 & 78.5 \\
          & ViT-B & -- & 78.3 & 78.2 & 78.4 \\
    \bottomrule 
    \end{tabular}
    }
\end{table}

% ================================================================== %
\section{Conclusion} \label{sec:conclusion}
% ================================================================== %
In this paper, we propose to utilize the feature extraction capabilities of VFMs as a simple and powerful solution for CI-FFREEDA. Thorough experiments comparing existing methods demonstrate that frozen VFMs effectively address domain gaps and class imbalances, providing a solution that overcomes the stagnation observed in existing approaches.
Our findings suggest that strong feature extractors, not complex adaptation or FL methods, are key to success in the real-world FL.

% ================================================================== %
% \section*{Acknowledgments}
% ================================================================== %
\vskip.5\baselineskip
\noindent
\textbf{Acknowledgments.}
This R\&D includes the results of "Research and development of optimized AI technology by secure data coordination (JPMI00316)" by the Ministry of Internal Affairs and Communications (MIC), Japan.
We thank the anonymous referees for their valuable comments.

% \section{REFERENCES}
% \label{sec:refs}

% List and number all bibliographical references at the end of the
% paper. The references can be numbered in alphabetic order or in
% order of appearance in the document. When referring to them in
% the text, type the corresponding reference number in square
% brackets as shown at the end of this sentence \cite{C2}. An
% additional final page (the fifth page, in most cases) is
% allowed, but must contain only references to the prior
% literature.

\clearpage
\vfill\pagebreak
% \bibliographystyle{IEEEbib_initial}
% \bibliography{_references_shrink.bib}

% ================================================================== %
% ================================================================== %
% ================================================================== %
\appendix
\section*{Appendix}
% ================================================================== %
% ================================================================== %
% ================================================================== %
\section{Detailed Experimental Setups}
\label{app:setup}
% ================================================================== %

\subsection{Detailed Setup for Sec.~\ref{subsec:setup}}

\noindent
\textbf{Data Partitioning and Evaluation.}
To experiment under CI-FFREEDA condition, we employ the following data partitioning strategy.
Initially, we select a source domain and randomly sample instances for each label to create an imbalanced source dataset, representing approximately 60\% of the entire dataset. From this subset, 80\% is allocated to the training set, while the remaining 20\% is designated as the validation set.
Subsequently, we select one target domain. For the test set, we allocate an equal number of samples for each label, comprising 20\% of the total dataset. The remaining samples are divided using Dirichlet sampling method, commonly employed in non-IID federated learning, to ensure class imbalance and heterogeneous distributions across clients. The samples assigned to each client are stratified to preserve the label distribution, with 80\% allocated to the training set and 20\% to the validation set.

In many existing domain adaptation studies that do not incorporate federated learning, data partitioning during target adaptation is not conducted, and models are typically evaluated using metrics derived from the training set after a fixed number of epochs. Contrary to their approach, our objective is to obtain a unified global model adapted to the target domain while avoiding overfitting to the training data held by each client. Consequently, we maintain the best global model on the server based on the evaluations conducted by each client using their validation set and measure MAR of the final global model using the test set from the administrator's perspective. Since the test set is balanced, MAR is equivalent to accuracy.

In the source training phase, 
three training/validation sets are constructed with different imbalance ratios. For each source imbalance ratio, experiments are conducted three times with different seed values, and the results of these nine runs are averaged. In the target adaptation phase, for each of the three source imbalance ratios, three different target imbalance ratios (representing imbalance distribution among clients) are applied, and the results from these nine runs are averaged.

\noindent
\textbf{Implementation Details.}
We adopt mini-batch SGD with momentum 0.9. The batch size is 64 and the learning rate is set to 0.01 for OH and 0.001 for VisDA. The learning rate of backbone during the training of ResNet is set to one-tenth of the default rate.
The feature dimension in bottleneck layer is set to 256. 
In ICPR, RandAugment~\cite{cubuk2020randaug} is used to generate augmented images. 
The number of augmentations to be given is set to 2 and the upper limit of augmentation magnitude is set to 9.
The degree of non-IID-ness of the data distribution, as determined by the Dirichlet sampling, is controlled by the parameter $\alpha$. Generally, non-IID-ness is more pronounced when $\alpha < 1$, resulting in one label being distributed preferentially to biased clients. In our experiments,  $\alpha$ is set to 0.5.

In experiments using ViT-S and ViT-B, the output of training samples processed by the ViT is stored in the feature bank for training, except during ICPR source training, ICPR target adaptation, and ISFDA target adaptation. Consequently, the data augmentation of random flip and random crop commonly employed in existing methods are not applied during training.

\subsection{Detailed Setup for Sec.~\ref{subsec:further_gap}}

In the further experiments in Sec.~\ref{subsec:further_gap}, three domain dataset in Office-Home; Clipart, Product, Real-World are split into three subset for source-set, target-set, and evaluation-set. In all scenarios, the evaluation-set is balanced. Meanwhile, the source-set is sampled with balanced label distribution in the source balance scenario (\textit{sb}), whereas random label distribution is applied in the source imbalance scenario (\textit{si}).
The target-set is further distributed among three clients, maintaining the label distribution in the target balance scenario (\textit{tb}) and applied a Dirichlet distribution in the target imbalance scenario (\textit{ti}).
This leads to the following four scenarios: source balance to target balance (\textit{sbtb}), source balance to target imbalance (\textit{sbti}), source imbalance to target balance (\textit{sitb}), and source imbalance to target imbalance (\textit{siti}). In Sec.~\ref{subsec:further_gap}, only the results of \textit{sbtb} and \textit{siti} are presented.
Note that in those experiments, the number of samples used for training in both the source and target domains is approximately half of that in Sec.~\ref{sec:experiments}, thus a direct comparison is not possible.

\subsection{Detailed Setup for Sec.~\ref{subsec:further_fed}}
In the experiments described herein, SHOT is employed as the base algorithm for SFDA. When integrated with FedProx, the regularization term is added to the existing SHOT loss function. The regularization factor is tuned from \{1.0, 0.1, 0.01, 0.001\}. A value of 0.001 is selected in VisDA with ResNet-101, while 0.1 is selected for the other settings. For FedETF, training is conducted using a dedicated bottleneck component and an ETF classifier. Following the official implementation, we replace the loss function in the self-training term with a balanced softmax loss, where the cross-entropy is corrected using the label distribution. Other implementation details are same as Sec.~\ref{sec:experiments}.

% ================================================================== %
\section{Complementary Results}
\label{app:result}
% ================================================================== %
\subsection{Complementary Results of Sec.~\ref{subsec:results}}

\begin{figure*}[]
    \centering
    \includegraphics[width=1.0\textwidth]{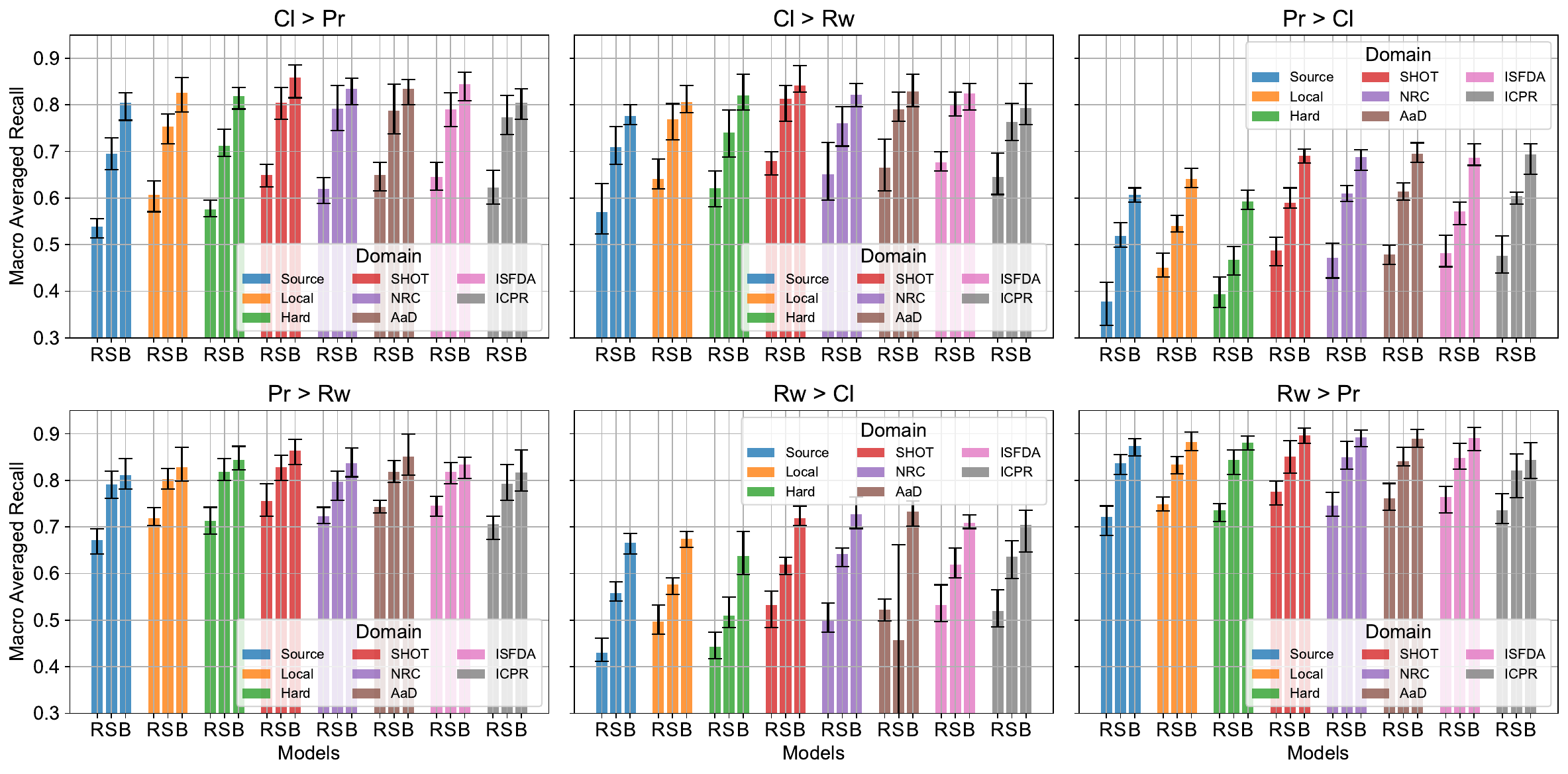}
    \caption{The whole results of federated target adaptation with OH and ResNet-50 (R), DINOv2 ViT-S (S), and ViT-B (B) conducted in Sec.~\ref{sec:experiments}. Panels and colors indicate different source-target domain pairs and SFDA methods, respectively. Each plot shows the average over nine runs, comprised of three different source imbalance ratios and three different target imbalance ratio. Error bars represent the maximum and minimum values from these nine runs. }
    \label{fig:app_tgt_oh}
\end{figure*}

The results of adaptation experiments for all domain patterns in OH are presented in Fig.~\ref{fig:app_tgt_oh}. In each domain-specific panel, different methods are represented by different colors. 
``Source'' refers to the baseline evaluation where the model, trained exclusively on the source domain, is directly applied to the target domain without any adaptation. ``Local'' represents the average performance across all clients, where each client independently trains its model without federated learning. ``Hard'' denotes a basic pseudo-labeling strategy: during local client training, the outputs of the current model are converted into a one-hot format and used as pseudo-labels.
Across nearly all methods, the accuracy follows the order: ViT-B (B), ViT-S (S), ResNet-50 (R). The accuracy among all methods is largely competitive, with no method significantly outperforming the others beyond the range of the error bars.

\subsection{Complementary Results of Sec.~\ref{subsec:further_gap}}
The whole results with TL and DA settings under \textit{sbtb}, \textit{sbti}, \textit{sitb}, and \textit{siti} scenarios are shown in Table~\ref{tab:app_whole_gap}. 
The source accuracy is averaged across three domains, three source sampling seeds, and three execution seeds, totaling 27 runs. The target accuracy is averaged across three domain pairs, three source sampling seeds, and three target distribution seeds in TL setting. In DA setting, it is averaged over six source-target pairs, three source sampling seeds, and three target distribution seeds, totaling 27 runs for TL and 54 runs for DA.

\begin{table}[ht]
    \centering
    \caption{The whole results conducted in Sec.\ref{subsec:further_gap} under transfer learning (TL) and domain adaptation (DA) settings considering domain gaps and label distribution gaps. The Scenario column represents the label distribution of the source (\textit{s}) and target (\textit{t}), balanced (\textit{b}) and imbalanced (\textit{i}). The decline in accuracy after the transfer or adaptation to the target (S2T diff.) is also presented.}
    \label{tab:app_whole_gap}
    \scalebox{0.75}{
    \begin{tabular}{c|cc|ccc}
    \toprule
            & Model  & Scenario  & Source & Target & S2T   \\
            &        &           & acc.   & acc.   & diff. \\
    \midrule
    \multirow{12}{*}{TL} & & \textit{sbtb} & 82.6 & 82.0 & -0.6 \\
     & \multirow{2}{*}{ResNet-50} &   \textit{sbti} & 82.6 & 81.0 & -1.6 \\
                 & & \textit{sitb} & 78.7 & 78.4 & -0.3 \\
                 & & \textit{siti} & 78.7 & 77.6 & -1.1 \\ \cline{2-6}
                 & & \textit{sbtb} & 87.4 & 86.4 & -1.0 \\
                 & \multirow{2}{*}{ViT-S} & \textit{sbti} & 87.4 & 86.0 & -1.4 \\ 
                 & & \textit{sitb} & 84.7 & 84.1 & -0.6 \\
                 & & \textit{siti} & 84.7 & 83.4 & -1.3 \\ \cline{2-6}
                 & & \textit{sbtb} & 89.9 & 90.1 & +0.2 \\
           & \multirow{2}{*}{ViT-B} & \textit{sbti} & 89.9 & 89.7 & -0.2 \\ 
                 & & \textit{sitb} & 88.8 & 87.9 & -0.9 \\ 
                 & & \textit{siti} & 88.8 & 87.2 & -1.6 \\
    \midrule
    \multirow{12}{*}{DA} & & \textit{sbtb} & 82.6 & 65.0 & -17.6 \\
     & \multirow{2}{*}{ResNet-50} &   \textit{sbti} & 82.6 & 64.0 & -18.6 \\
                 & & \textit{sitb} & 78.7 & 62.0 & -16.7 \\
                 & & \textit{siti} & 78.7 & 60.8 & -17.9 \\ \cline{2-6}
                 & & \textit{sbtb} & 87.4 & 76.8 & -10.6 \\
                 & \multirow{2}{*}{ViT-S} & \textit{sbti} & 87.4 & 74.2 & -13.2 \\
                 & & \textit{sitb} & 84.7 & 73.1 & -11.6 \\
                 & & \textit{siti} & 84.7 & 71.4 & -13.3 \\ \cline{2-6}
                 & & \textit{sbtb} & 89.9 & 82.6 & -7.3  \\
           & \multirow{2}{*}{ViT-B} & \textit{sbti} & 89.9 & 80.8 & -9.1  \\
                 & & \textit{sitb} & 88.8 & 79.4 & -9.4 \\
                 & & \textit{siti} & 88.8 & 78.1 & -10.7 \\
    \bottomrule 
    
    \end{tabular}
    }
\end{table}

\subsection{Complementary Results of Sec.~\ref{subsec:further_fed}}

\begin{figure}[h]
    \centering
    \includegraphics[width=0.48\textwidth]{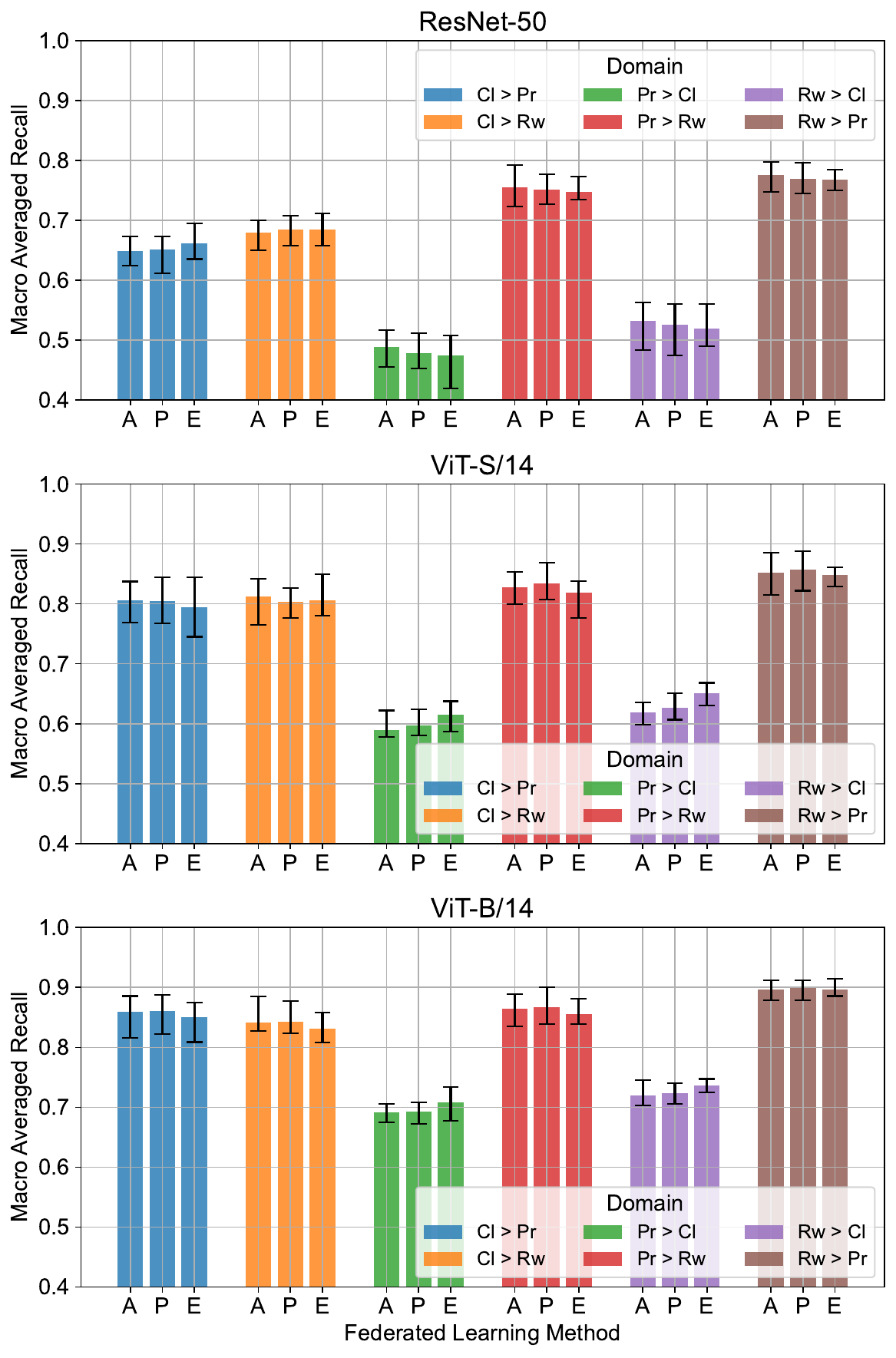}
    \caption{The comparison among three different federated learning method, FedAvg (A), FedProx (P), and FedETF (E), with OH. Note that in this figure, colors indicate different source-target pairs. Each plot shows the average over nine runs, comprised of three different source imbalance ratios and three different target imbalance ratio. Error bars represent the maximum and minimum values from these nine runs.}
    \label{fig:app_oh_fed}
\end{figure}

\begin{figure}[h]
    \centering
    \includegraphics[width=0.48\textwidth]{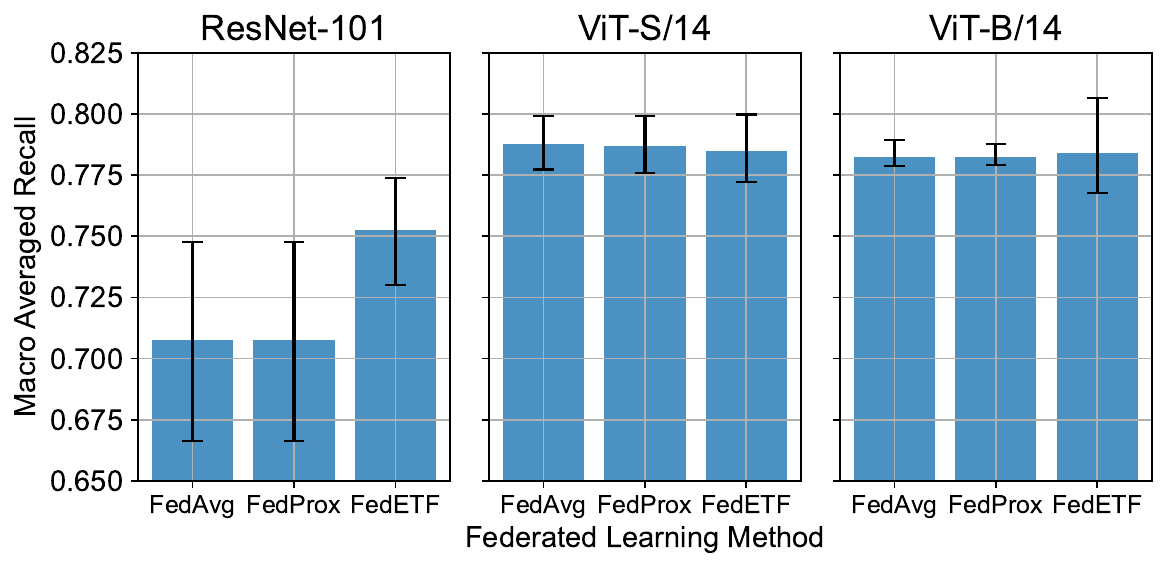}
    \caption{The comparison among three different federated learning method, FedAvg, FedProx, and FedETF, with VisDA. Details of the plot are identical to those in Fig.~\ref{fig:app_oh_fed}.}
    \label{fig:app_visda_fed}
\end{figure}

Fig.~\ref{fig:app_oh_fed} illustrates the error bars with different seeds for OH dataset, as indicated Table~\ref{tab:further_icls}. Fig.~\ref{fig:app_visda_fed} is the same figure for VisDA.

% ================================================================== %
\section{Other Models and Backbone Fine-tuning}
\label{app:other}
% ================================================================== %

In edge environments, such as a single client participating in federated learning, models like ViT-S/14 (21M parameters) and ViT-B/14 (86M), mainly used in this study, may not be sufficiently lightweight for efficient processing. To investigate more resource-efficient alternatives, we include in our evaluation lighter models with available pretrained weights: LightViT-Tiny~\cite{huang2022lightvit} (9.4M), TinyViT-5M~\cite{wu2022tinyvit} (5.4M), and the image encoder ViT-8M/16 of TinyCLIP~\cite{wu2023tinyclip} (8M). The results are shown for cases where these models are used as frozen backbones, as well as when the pretrained weights are used for initialization and the entire backbone is fine-tuned.
Furthermore, the results also include the performance of DINOv2 when the backbone is fine-tuned, as well as the results using the latest models released by Meta: DINOv3~\cite{simeoni2025dinov3}, ViT-S/16 (21M), ViT-S+/16 (28M), and ViT-B/16 (86M), used as frozen backbones.
For each setting, we conducted a parameter search over learning rates from \{0.1, 0.01, 0.001\}, weight decay values \{0.01, 0.001, 0.0001\}, and backbone learning rate ratios during fine-tuning \{0.1, 0.01, 0.001, 0.0001\}. The best-performing results from these configurations with SHOT and FedAvg are reported.

\begin{figure*}[h]
    \centering
    \includegraphics[width=1.0\textwidth]{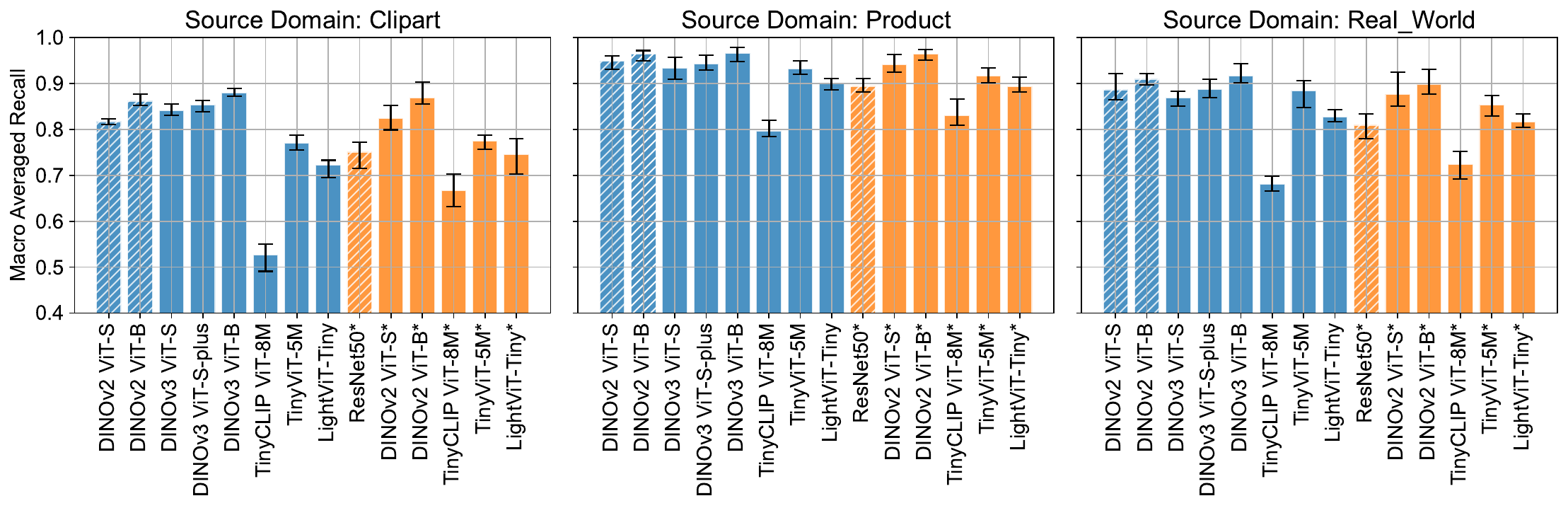}
    \caption{Accuracy of the source domain on the OH dataset with various models. Blue bars indicate results obtained using a frozen backbone, while orange bars, denoted by an asterisk (*) following the model name, represent results with a fine-tuned backbone. Hatched bars correspond to the original models discussed in this paper. Each plot is the average over nine runs, comprised of three different source imbalance ratios and three execution seeds for each source sampling. Error bars represent the maximum and minimum values of nine runs.}
    \label{fig:app_ts_oh}
\end{figure*}

\begin{figure}[h]
    \centering
    \includegraphics[width=0.48\textwidth]{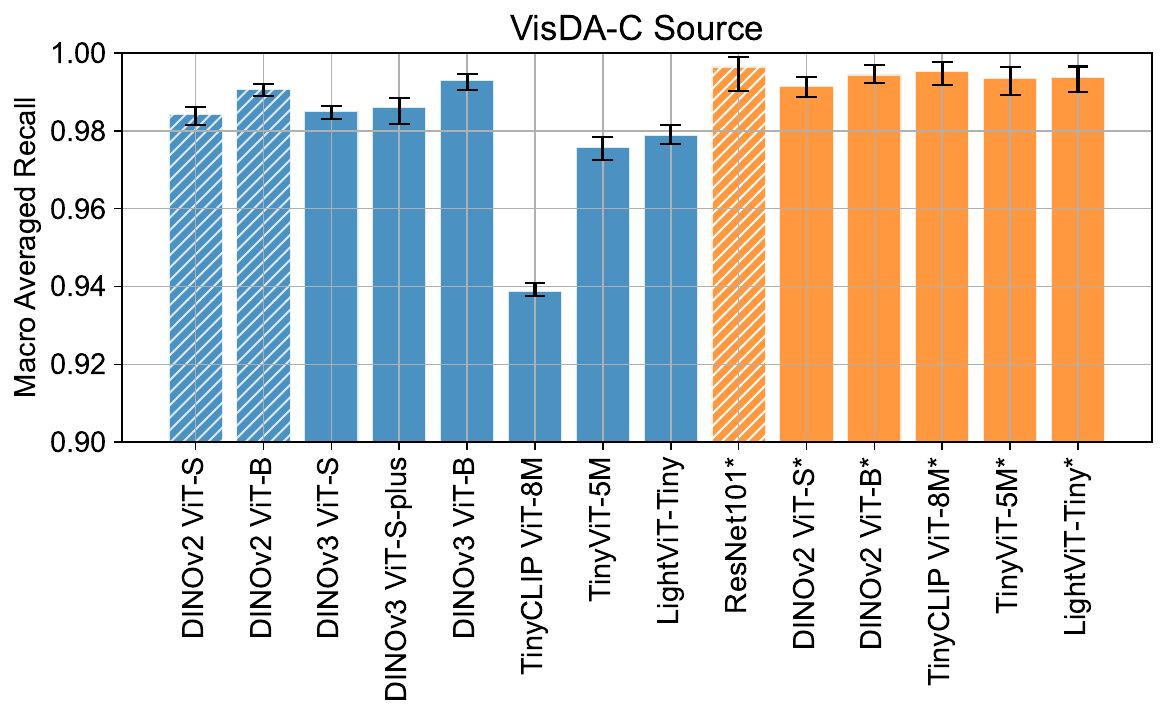}
    \caption{Accuracy of the source domain on the VisDA dataset with various models. Details are same as Fig.~\ref{fig:app_ts_oh}.}
    \label{fig:app_ts_visda}
\end{figure}

\noindent
\textbf{Source Training Accuracy.}
The training results on the source domain show that TinyViT achieves comparable to DINOv2 ViT-S among lightweight models in OH (Fig.~\ref{fig:app_ts_oh}), while both LightViT and TinyViT demonstrate high accuracy on VisDA (Fig.~\ref{fig:app_ts_visda}). When fine-tuning lightweight models, the source accuracy remains nearly the same as in the frozen case on OH, whereas higher accuracy is observed in VisDA.

\begin{figure*}[h]
    \centering
    \includegraphics[width=1.0\textwidth]{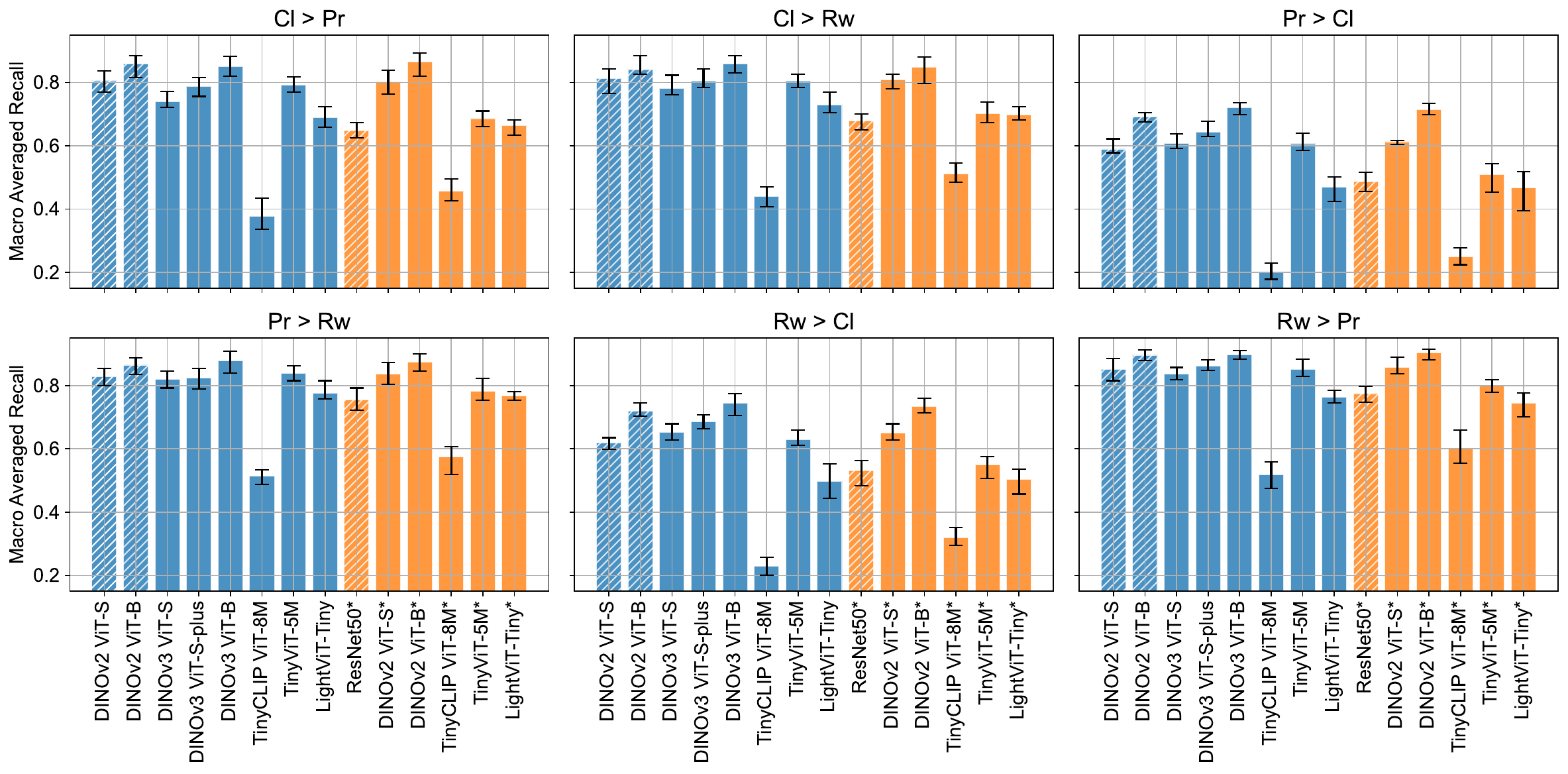}
    \caption{The whole results of federated target adaptation on the OH dataset with various models. Details of the bars are same as Fig.~\ref{fig:app_ts_oh}. Each plot shows the average over nine runs, comprised of three different source imbalance ratios and three different target imbalance ratio. Error bars represent the maximum and minimum values from these nine runs.}
    \label{fig:app_tt_oh}
\end{figure*}

\begin{figure}[h]
    \centering
    \includegraphics[width=0.48\textwidth]{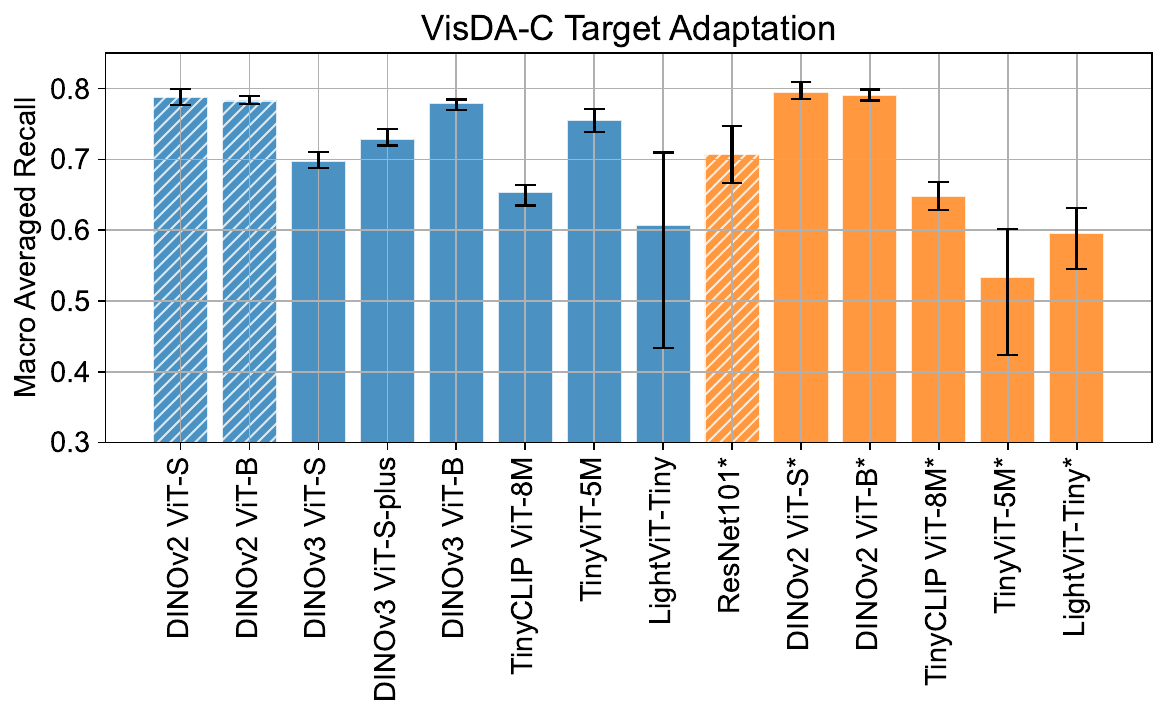}
    \caption{Results of federated target adaptation on the VisDA dataset with various models. Details are same as Fig.~\ref{fig:app_tt_oh}.}
    \label{fig:app_tt_visda}
\end{figure}

\noindent
\textbf{Target Adaptation Accuracy.}
After target adaptation, the frozen TinyViT achieves performance nearly equivalent to DINOv2 ViT-S in OH (Fig.~\ref{fig:app_tt_oh}) and slightly lower in VisDA (Fig.~\ref{fig:app_tt_visda}), yet in both cases it outperforms the fine-tuned ResNet. On the other hand, fine-tuning lightweight models results in decreased accuracy, suggesting that more sensitive adjustment may be necessary. Fine-tuning DINOv2 yields accuracy comparable to or slightly better than its frozen counterpart, but considering computational costs, using the frozen backbone proves to be advantageous.

% ================================================================== %
\section{Computational and Communication Costs}
\label{app:cost}
% ================================================================== %

The proposed method in this study reduces computational resource consumption during backpropagation by freezing the VFM component. Table~\ref{tab:costs} shows the number of FLOPs and the size of models that must be transferred between the server and the client. FLOPs are computed using the calflops\footnote{\href{https://github.com/MrYxJ/calculate-flops.pytorch}{https://github.com/MrYxJ/calculate-flops.pytorch}} library. Since batch normalization is used, the number of batch is calculated as 2 and the result is halved to obtain the FLOPs per image. Additionally, when fine-tuning, the computational cost of the backward pass is assumed to be twice that of the forward pass. Model sizes are based on the actual saved size using the standard method of PyTorch.

\begin{table}[ht]
    \centering
    \caption{FLOPs and model sizes for each model and training strategy.}
    \label{tab:costs}
    \scalebox{0.75}{
    \begin{tabular}{cc|cc}
    \toprule
    Method & Model & FLOPs & Model Size \\
    \midrule
    \multirow{2}{*}{Fine-tuning}        & ResNet-50  & 24.6 G & 94 MB  \\
                                        & ResNet-101 & 46.9 G & 169 MB \\ \cline{1-4}
    \multirow{2}{*}{Frozen VFMs}        & ViT-S      & 11.0 G & $<$ 1 MB \\
                                        & ViT-B      & 43.9 G & $<$ 1 MB \\ \cline{1-4}
    Frozen VFMs                         & ViT-S      & 0.61 M & $<$ 1 MB \\
    with backbone skipped               & ViT-B      & 1.20 M & $<$ 1 MB \\
    \midrule
    \bottomrule
    \end{tabular}
    }
\end{table}
    
In the case of ResNet fine-tuning, training requires approximately three times the computational cost of a single forward pass. The resulting trained model, which is roughly 100~MB in size, must be transmitted. In contrast, when using frozen VFMs, the computational cost for ViT-S is reduced by half compared to ResNet-50. Moreover, only the bottleneck and classifier components, which together total less than 1~MB, need to be transmitted to the server, significantly improving communication efficiency. Furthermore, if the outputs of the frozen VFMs is stored in a feature bank during the initial training phase and the backbone computation is skipped thereafter, enabling training only of the bottleneck and classifier (Frozen VFMs with backbone skipped), it is possible to entirely eliminate backbone computation during training.

% ================================================================== %

\end{document}